\documentclass{svproc}

\usepackage{url}
\usepackage{helvet}         
\usepackage{courier}        
\usepackage{color}
\usepackage{makeidx}         
\usepackage{graphicx}        
\usepackage[misc]{ifsym}
\usepackage{multicol}        
\usepackage[bottom]{footmisc}
\usepackage{tabularx}
\usepackage{multirow}
\usepackage{natbib}
\usepackage{algorithm}
\usepackage{algorithmic}
\usepackage[center]{caption}
\usepackage{longtable}
\usepackage{rotating}
\usepackage{amsmath}
\usepackage{amssymb}
\usepackage{subfigure}
\usepackage{placeins}

\begin{document}
\mainmatter              
\title{Rethinking LLM-Guided Visual Question Answering: Does Deeper Knowledge Integration Really Help?}
\titlerunning{Rethinking LLM-Guided Visual Question Answering}
\author{Aneerban Saha\inst{1,*} \and Vijay Kumar\inst{2} \and Sameer Madhav D.\inst{3} \and Priyam Srivastava\inst{2} \and Dhruv Bansal\inst{2}}
\authorrunning{Aneerban Saha et al.}
\tocauthor{Aneerban Saha, Vijay Kumar, Sameer Madhav D., Priyam Srivastava, and Dhruv Bansal}
\institute{Independent Researcher, West Bengal, India\\
\email{aneerbansaha22@gmail.com}$^{*}$
\and
Independent Researchers, Delhi, India\\
\email{kumarvijay2003.vk@gmail.com, srivastavapriyam652@gmail.com,}\\
\email{dhruvbansal2804@gmail.com}
\and
Independent Researcher, Karnataka, India\\
\email{sameermadhav.5@gmail.com}\\[6pt]
$^{*}$Corresponding author: Aneerban Saha, \email{aneerbansaha22@gmail.com}}

\maketitle              

\begin{abstract}
Some questions about an image need commonsense or factual knowledge that its pixels do not contain. Large Language Models (LLMs) can supply this knowledge to Visual Question Answering (VQA) systems, but the depth at which their guidance should enter a vision--language model remains unclear. We prompt the 70-billion-parameter Llama~3.3 model to describe, for each question, the visual concepts to focus on, the distractors to ignore and the external facts needed. We inject this guidance into a frozen Vision Transformer (ViT) paired with an instruction-tuned Text-to-Text Transfer Transformer (T5) in two ways: decision-level late fusion and attention-matrix injection. On the Outside Knowledge VQA (OK-VQA) benchmark, and contrary to the common view that deeper is better, the shallow route wins: late fusion reaches 39.4\% exact match, +10.0 points above an unguided baseline, while attention injection reaches only 31.8\% under data-scarce fine-tuning. An LLM paraphrasing pipeline also enlarges the training set more than tenfold. The results suggest that integration depth should be matched to the data regime.
\end{abstract}

\section{Introduction}\label{sec:intro}
Visual Question Answering (VQA) sits where vision meets language: given an image and a free-form question, a system has to return a correct answer. Standard VQA models cope well when the answer is visible, but they struggle as soon as commonsense or factual knowledge is required. Consider the question \emph{``How old do you have to be in Canada to do this?''} asked about a photograph of adults drinking wine; no inspection of the pixels reveals the legal drinking age. The Outside Knowledge VQA (OK-VQA) benchmark~\cite{marino2019okvqa} was built to probe exactly this gap, since each of its questions depends on knowledge that lies outside the image.

Several recent studies turn to large language models (LLMs) as on-demand sources of such knowledge \cite{yang2022pica,gui2022kat,shao2023prophet}. The usual recipe, however, treats the LLM as a text-only oracle that produces the answer itself. Approaches that instead build knowledge into a trainable architecture tend to equate more elaborate fusion with better accuracy \cite{he2026kvkbvqa,dong2024mail}, even though evidence from other multimodal settings shows that deep or early fusion can overfit when labelled data are limited and may lose to simpler, decision-level designs. This tension motivates our central question: \emph{given structured guidance produced by an LLM, at what depth should it enter a multimodal transformer, and does deeper injection actually help?}

Concretely, we prompt Llama~3.3 with 70 billion parameters (Llama~3.3~70B) to emit, for every OK-VQA sample, a structured reasoning object with three fields: \texttt{visual\_focus} (objects and attributes relevant to the question), \texttt{ignore} (distractors), and \texttt{knowledge} (external facts needed for the answer). A frozen Contrastive Language--Image Pre-training (CLIP) text encoder maps these fields to a dense guidance vector. The vector enters a shared model that pairs a Vision Transformer (ViT) encoder with a FLAN-T5 answer generator, that is, a Text-to-Text Transfer Transformer (T5) instruction-tuned on the Finetuned Language Net (FLAN) collection. We compare two routes: a shallow, decision-level \emph{late fusion} path, and a deeper \emph{attention-matrix injection} that modifies the query and key projections inside the vision encoder.

Our contributions are: (1)~a pipeline that turns LLM reasoning into structured, machine-readable guidance consumed by a compact trainable VQA model, instead of relying on the LLM to produce each answer; (2)~a controlled comparison of shallow and deep knowledge integration under identical data-scarce fine-tuning on OK-VQA; (3)~a counter-intuitive empirical result, namely that shallow late fusion beats deep attention modification on every metric (exact match 39.4 vs.\ 31.8; token F1 41.3 vs.\ 33.3); and (4)~an LLM-based augmentation pipeline that grows the filtered training set from 3{,}144 to roughly 36{,}000 samples.

Section~\ref{sec:related} surveys prior work, and Section~\ref{sec:dataset} explains how the dataset and the guidance were built. Section~\ref{sec:arch} describes the model variants, Section~\ref{sec:results} presents and analyses the experiments, and Section~\ref{sec:con} closes the paper.

\section{Related Work}\label{sec:related}

\subsection{Knowledge-Based Visual Question Answering}
Marino et al.~\cite{marino2019okvqa} released OK-VQA, a set of 14{,}055 questions on Microsoft Common Objects in Context (MS~COCO) images that cannot be answered without world knowledge, and showed a steep drop in accuracy for models designed for visually grounded questions. Later work tried to close this gap through retrieval. Knowledge Reasoning with Implicit and Symbolic rePresentations (KRISP)~\cite{marino2021krisp} combined implicit knowledge learned during multimodal pretraining with symbolic reasoning over knowledge graphs, and Multi-modal Answer Validation using External knowledge (MAVEx)~\cite{wu2022mavex} checked candidate answers against evidence drawn from Wikipedia, ConceptNet and web images.

The field changed markedly once LLMs such as Generative Pre-trained Transformer~3 (GPT-3) began to act as on-the-fly knowledge sources. Prompting GPT-3 via the use of Image Captions (PICa)~\cite{yang2022pica} fed captions and a handful of solved examples to GPT-3, and with only 16 in-context examples it surpassed the best supervised OK-VQA result of its time. The Knowledge Augmented Transformer (KAT)~\cite{gui2022kat} merged GPT-3 outputs with facts retrieved from Wikidata inside a single encoder--decoder, while Prophet~\cite{shao2023prophet} used a lightweight VQA model to propose candidate answers and examples that sharpen the GPT-3 prompt. On the fusion side, Modality-Aware Integration with LLMs (MAIL)~\cite{dong2024mail} derives paired scene and concept graphs through two-stage LLM prompting and merges them with a dual-branch graph fusion module. KV-KBVQA~\cite{he2026kvkbvqa}, a Kolmogorov--Arnold Network (KAN) based ViT for knowledge-based VQA, replaces standard layers with KAN units and adds bidirectional residual cross-modal attention, attributing earlier shortcomings to overly simple attention designs.

Two patterns stand out. Prompting-based systems (PICa, Prophet) rely on a billion-parameter LLM to produce every answer, which makes inference costly and hard to inspect. Systems that embed knowledge in a trainable model (KAT, MAIL, KV-KBVQA) keep adopting more intricate fusion, implicitly treating depth of integration as a route to accuracy; this is the assumption our experiments test.

\subsection{Vision-Language Fusion Strategies}
Vision--language architectures have moved from dual-stream pipelines over region features to unified transformers that operate on image patches. Learning Cross-Modality Encoder Representations from Transformers (LXMERT)~\cite{tan2019lxmert} linked separate visual and language encoders through a dedicated cross-modality encoder, whereas the Vision-and-Language Transformer (ViLT)~\cite{kim2021vilt} dropped convolutional feature extraction and fed image patches and word tokens into one transformer. Shin and Narihira~\cite{shin2021transformer} explored a minimal variant in which independently encoded modalities meet only in a joint embedding space, a design close in spirit to our late-fusion route. Controlled comparisons on red--green--blue plus depth (RGB-D) object recognition further show that late fusion often outperforms earlier fusion in vision transformers~\cite{tziafas2022fusion}. The best fusion depth therefore appears to depend on the amount of labelled data and on the strength of the pretrained unimodal encoders, which is the trade-off our study examines for LLM-derived guidance.

\subsection{Attention Guidance and Modification in Transformers}
Another line of work steers transformer behaviour by manipulating attention. Samaran et al.~\cite{samaran2021attending} supervised attention maps directly with grounding signals and found that pushing attention toward annotated regions does not reliably improve task accuracy, an early cautionary result. In text-only models, the Post-hoc Attention STeering Approach (PASTA)~\cite{zhang2024pasta} rescales a subset of attention heads to emphasise tokens that the user marks as important. Both studies confirm that attention is a workable injection point, but neither asks whether the benefit survives when the injected signal comes from an LLM and labelled examples are scarce, which is the setting we target.

\subsection{Foundation Models and LLM-Based Augmentation}
Our pipeline chains several pretrained components. A Vision Transformer~\cite{dosovitskiy2021vit} encodes each image as a sequence of patches, a CLIP text encoder~\cite{radford2021clip} turns guidance phrases into embeddings aligned with visual concepts, and Bootstrapping Language-Image Pre-training (BLIP)~\cite{li2022blip} supplies image captions. Answers are generated by FLAN-T5~\cite{chung2024flant5}, the instruction-tuned successor of T5~\cite{raffel2020t5}, in a text-to-text formulation. For data augmentation, AugGPT~\cite{dai2024auggpt} showed that rephrasing training sentences with ChatGPT improves few-shot text classification; we adapt this idea to enlarge our question set.

\textbf{Research gap.} Existing methods either call an LLM to produce each answer or build knowledge into increasingly complex fusion layers. What is missing is a controlled, head-to-head comparison of shallow and deep integration that uses the \emph{same} LLM-generated structured guidance, the \emph{same} frozen backbone and the same training budget on OK-VQA.

\section{Dataset Preparation and Guidance Generation}\label{sec:dataset}

\subsection{OK-VQA Preprocessing}
Every OK-VQA sample pairs an MS~COCO photograph with a question and several answers written by human annotators. Because annotators differ in spelling, case and phrasing, each answer is first normalised (lowercased, stripped of punctuation, and cleared of redundant whitespace), and a single training target is then chosen by majority vote, as shown in Eq.~\ref{eq:majority}.

\begin{equation}\label{eq:majority}
a^{*} = \arg\max_{a \in \mathcal{A}_q} \; \mathrm{count}(a)
\end{equation}

here, $\mathcal{A}_q$ denotes the multiset of annotated answers for question $q$; for instance, $[\textrm{``dog''},\textrm{``a dog''},\textrm{``dog''},\textrm{``puppy''},\textrm{``dog''}] \rightarrow \textrm{``dog''}$.

Since the answer distribution has a long tail, only samples whose target falls among the $K=500$ most frequent answers are kept. This leaves 3{,}144 training samples spread over 500 answer classes, roughly six per class, a severely data-scarce regime that motivates both our augmentation pipeline and the central research question. Each image is rescaled to $224\times224$ pixels and normalised with the preprocessing that accompanies the pretrained ViT.

\subsection{LLM-Based Reasoning Generation}
OK-VQA provides no reasoning annotations, so we construct them automatically. A BLIP captioner first describes each image in one sentence $c$; Llama~3.3~70B then receives the question $q$ together with $c$ and returns a JavaScript Object Notation (JSON) object of the form given in Eq.~\ref{eq:guidance}.

\begin{equation}\label{eq:guidance}
G = \{\, V, \; I, \; W \,\}
\end{equation}

here, $V = \{v_1,\dots,v_m\}$ collects the \emph{visual focus} concepts, $I$ the elements to \emph{ignore}, and $W = \{w_1,\dots,w_n\}$ the external \emph{knowledge} statements. For the wine example above, the LLM returns $V = \{\textrm{man},\textrm{woman},\textrm{wine},\dots\}$, $I = \{\textrm{party decorations}\}$ and $W = \{\textrm{legal drinking age in Canada},\dots\}$. The reasoning objects are produced in one offline pass at temperature $0.2$ and stored, so the LLM is queried once per sample rather than repeatedly during training; for a previously unseen image--question pair, one captioning pass and one LLM call are still required before the model can be applied.

\subsection{LLM-Based Data Augmentation and Training Instances}
To counter the shortage of about six samples per class, the same LLM paraphrases each question under three constraints: the image stays the same, the answer stays the same, and only the wording of the question changes (e.g., \emph{``How old do you have to be in Canada to do this?''} $\rightarrow$ \emph{``What is the legal drinking age in Canada?''}). The pipeline uses shard-based caching, crash recovery and duplicate-free merging, and it grows the dataset from 3{,}144 to approximately 36{,}000 samples, following the paraphrase-based augmentation idea of AugGPT~\cite{dai2024auggpt}.

Inputs follow the T5 text-to-text convention, with the guidance context formed by concatenating the focus and knowledge strings, as in Eq.~\ref{eq:input}.

\begin{equation}\label{eq:input}
x_{\mathrm{text}} = \texttt{question: } q \;\; \texttt{context: } \big(V \,\Vert\, W\big)
\end{equation}

here, the input string is tokenised into token identifiers and an attention mask $(\mathbf{x}_{\mathrm{ids}}, \mathbf{x}_{\mathrm{mask}})$ with a maximum length of 64 tokens, and each target answer is tokenised into decoder labels $\mathbf{y}$ with a maximum length of 8 tokens.

\section{Model Architectures}\label{sec:arch}

\subsection{Shared Backbone}
\FloatBarrier
\begin{figure}[!htbp]
\centering
\includegraphics[width=0.60\textwidth]{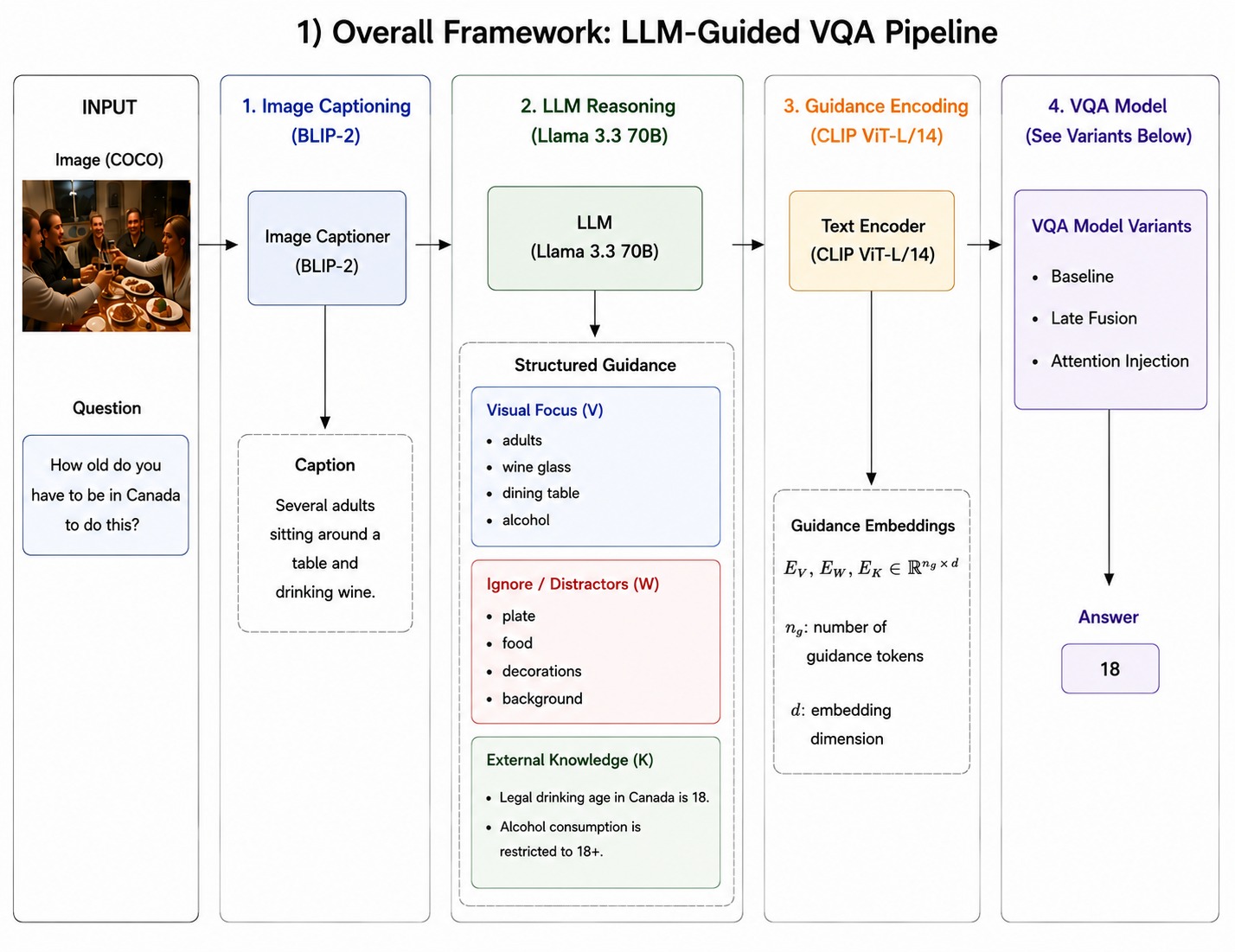}
\caption{Overall LLM-guided VQA framework: caption generation, structured reasoning, guidance encoding, and downstream model variants.}
\label{fig:backbone}
\end{figure}
\FloatBarrier
All variants share one architecture, so only the way knowledge is integrated changes between them (Fig.~\ref{fig:backbone}). Images are encoded by a pretrained ViT-Base model~\cite{dosovitskiy2021vit}, and its classification ([CLS]) token embedding $\mathbf{z}_{\mathrm{cls}}\in\mathbb{R}^{768}$ serves as the global visual representation. Questions are processed by FLAN-T5~\cite{chung2024flant5}. The structured guidance (visual focus and external knowledge) is embedded with the frozen CLIP text encoder~\cite{radford2021clip}, mean-pooled, and projected into the ViT feature space according to Eq.~\ref{eq:proj}.

\begin{equation}\label{eq:proj}
\mathbf{P}=\mathbf{W}_{p}\mathbf{P}_{\mathrm{raw}}+\mathbf{b}_{p}
\end{equation}

here, $\mathbf{P}_{\mathrm{raw}} \in \mathbb{R}^{512}$ is the mean-pooled CLIP embedding, $\mathbf{W}_{p}$ and $\mathbf{b}_{p}$ are learnable projection parameters, and $\mathbf{P} \in \mathbb{R}^{768}$ is the guidance vector shared by both integration strategies below.

\subsection{Baseline (No Guidance)}

\FloatBarrier
\begin{figure}[!htbp]
\centering
\includegraphics[width=0.60\textwidth]{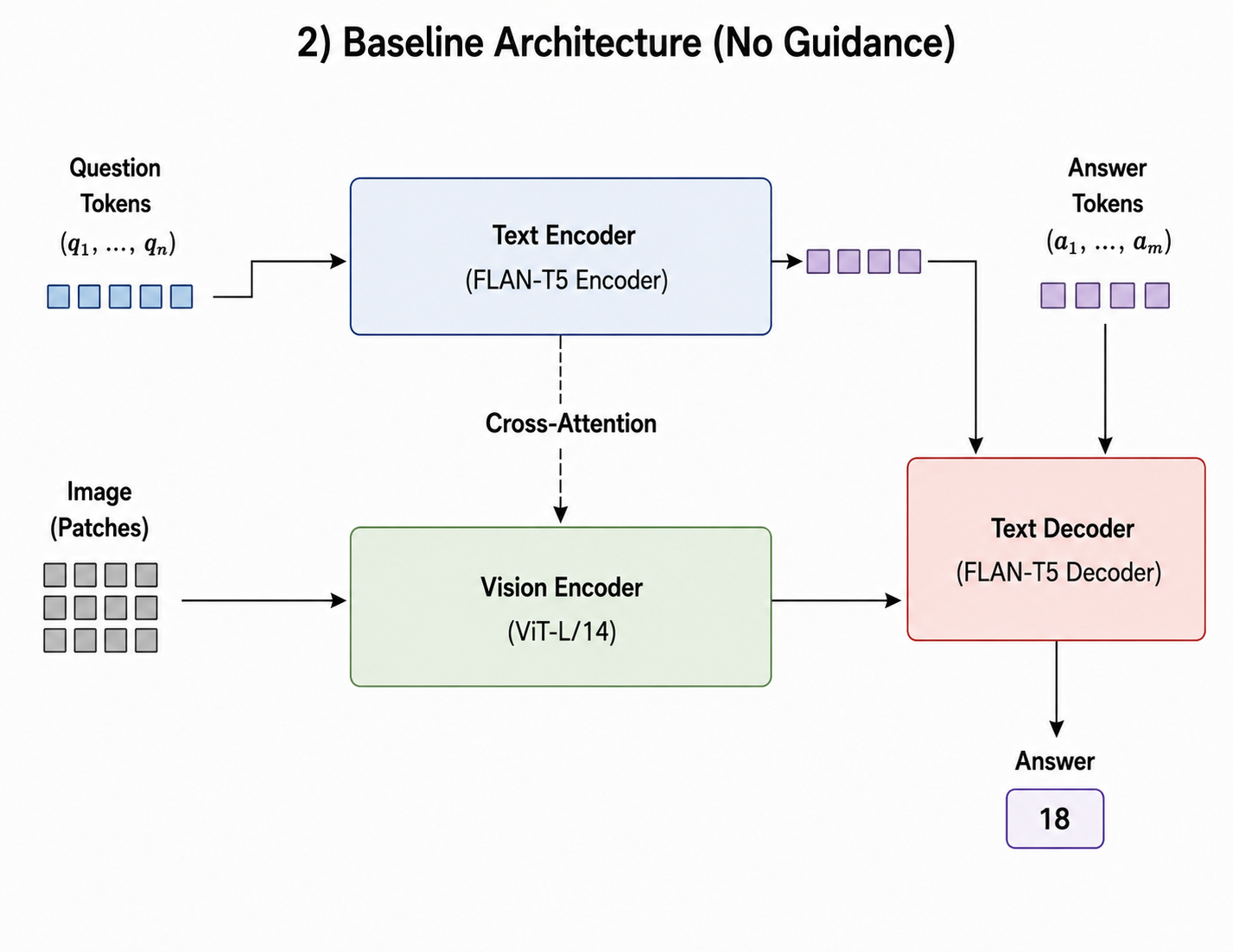}
\caption{Baseline VQA architecture: ViT image features and FLAN-T5 question embeddings are fused via cross-attention before decoding.}
\label{fig:baseline}
\end{figure}
\FloatBarrier

The baseline sends only visual information to the answer generator (Fig.~\ref{fig:baseline}). The [CLS] embedding is projected to the T5 embedding size and placed in front of the question tokens as a single prefix token, as written in Eq.~\ref{eq:baseline}.

\begin{equation}\label{eq:baseline}
\mathbf{h}_{0} = \mathbf{W}_{v}\,\mathbf{z}_{\mathrm{cls}}, \qquad
\tilde{\mathbf{E}} = [\,\mathbf{h}_{0}\,;\,\mathbf{E}(\mathbf{x}_{\mathrm{ids}})\,]
\end{equation}

here, $\mathbf{E}(\cdot)$ is the T5 input embedding layer and $[\,\cdot\,;\,\cdot\,]$ denotes concatenation along the sequence. Training minimises the usual sequence-to-sequence cross-entropy in Eq.~\ref{eq:loss}.

\begin{equation}\label{eq:loss}
\mathcal{L} = -\sum_{k=1}^{|\mathbf{y}|} \log p_{\theta}\!\left(y_{k} \mid y_{<k}, \tilde{\mathbf{E}}\right)
\end{equation}

here, $y_{k}$ is the $k$-th answer token and $\theta$ collects the trainable parameters.

\subsection{Decision-Level Late Fusion}
\begin{figure}[!htbp]
\centering
\includegraphics[width=0.60\textwidth]{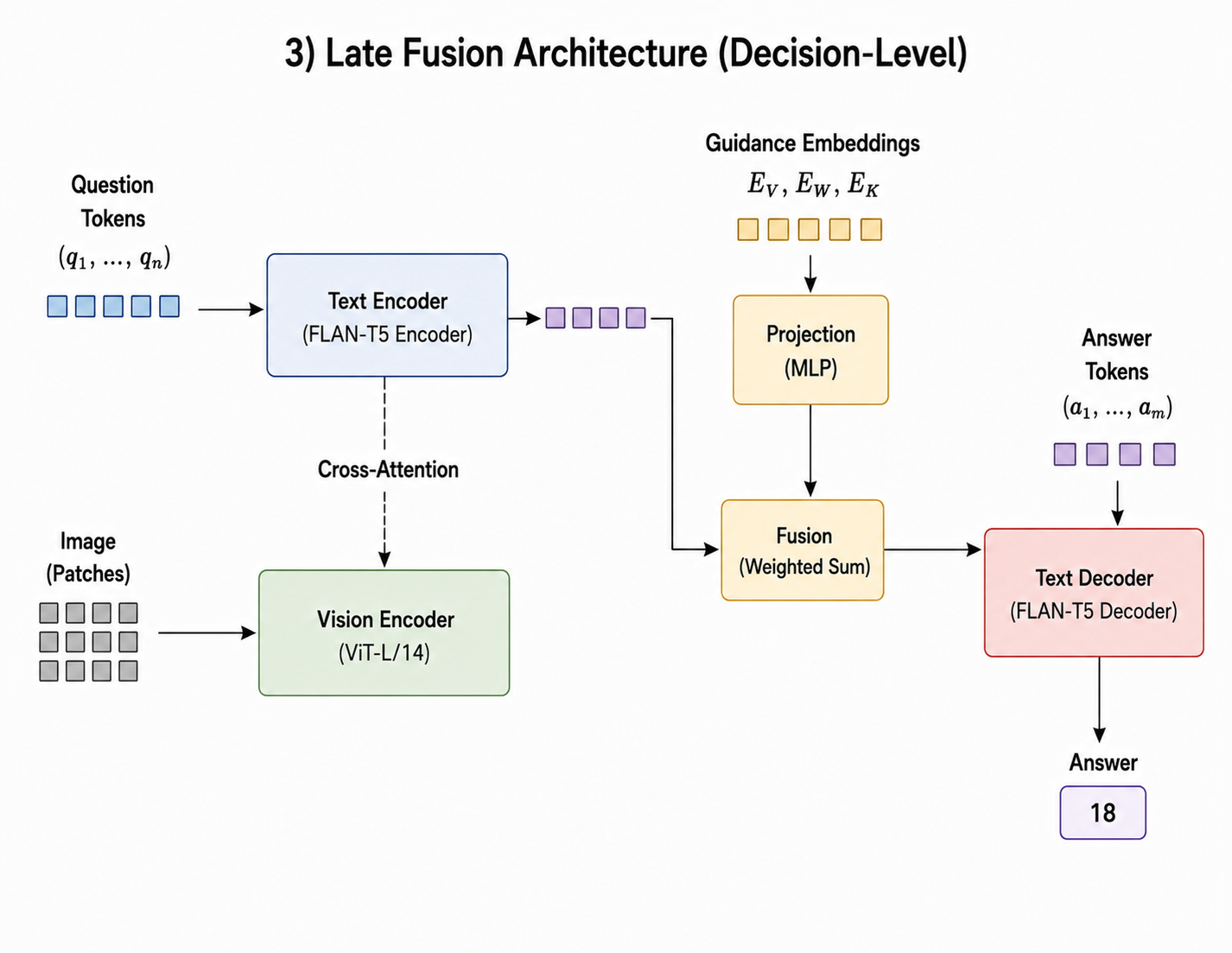}
\caption{Late fusion: projected guidance is fused with the global visual representation immediately before decoding.}
\label{fig:late_fusion}
\end{figure}

In late fusion, guidance meets the visual features only at the last stage before decoding (Fig.~\ref{fig:late_fusion}). The [CLS] embedding and the guidance vector are concatenated and passed through a fusion layer, as defined in Eq.~\ref{eq:fusion}.

\begin{equation}\label{eq:fusion}
\mathbf{f} = \mathbf{W}_{f}\,[\,\mathbf{z}_{\mathrm{cls}} \,\Vert\, \mathbf{P}\,] + \mathbf{b}_{f}, \qquad
\mathbf{h}_{0} = \mathbf{W}_{v}\,\mathbf{f}
\end{equation}

here, $[\,\cdot \,\Vert\, \cdot\,]$ denotes vector concatenation, mapping $\mathbb{R}^{1536}$ to the T5 embedding size $d_{T5}$. The fused vector is prepended to the T5 input exactly as in the baseline, so the guidance shapes the final decision but never the extraction of visual features; this is the shallowest form of integration.
\FloatBarrier
\FloatBarrier

\subsection{Deep Integration via Attention-Matrix Injection}
\begin{figure}[!htbp]
\centering
\includegraphics[width=0.60\textwidth]{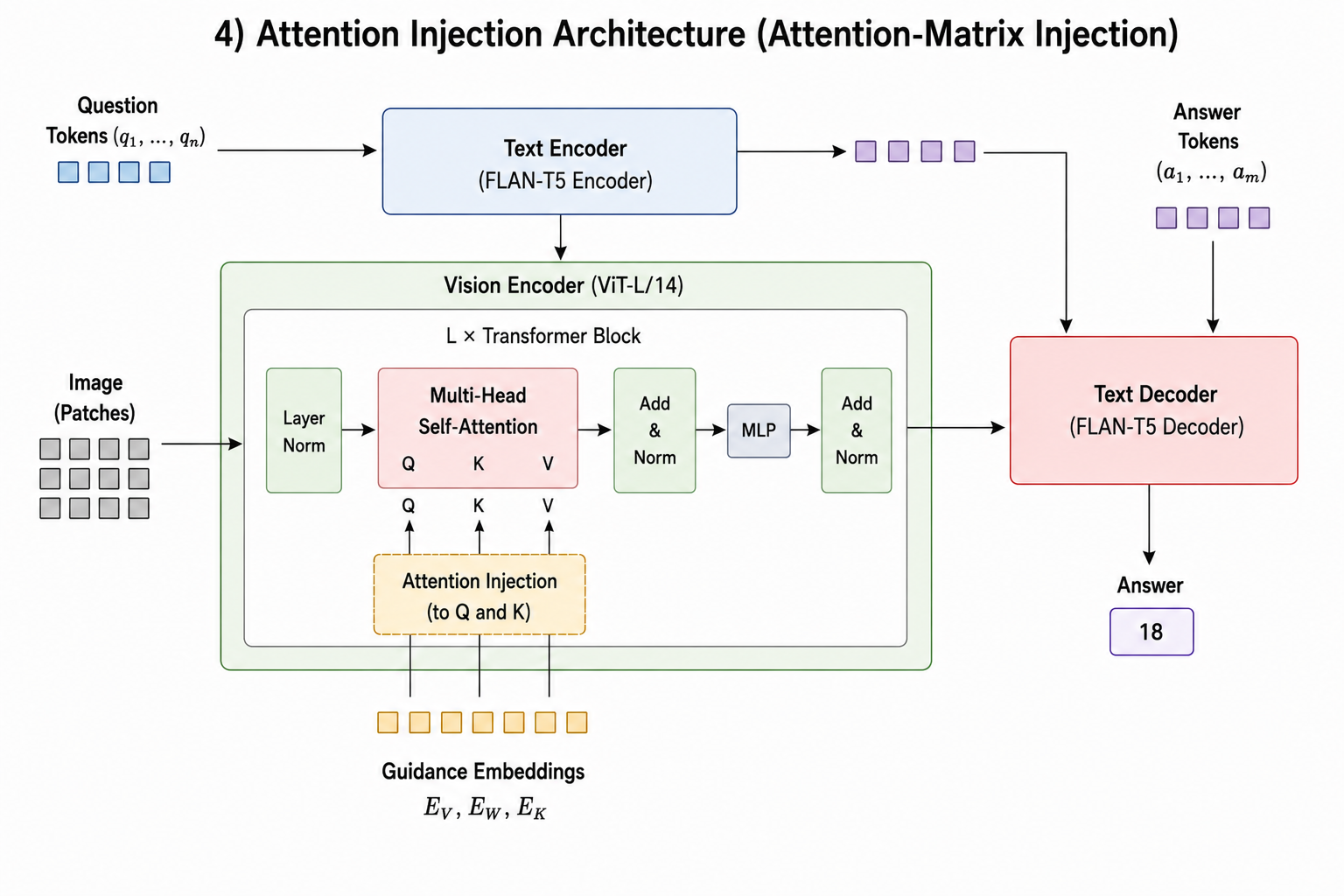}
\caption{Attention injection: projected guidance modifies the query and key projections inside the vision transformer.}
\label{fig:attention_injection}
\end{figure}

This variant feeds guidance directly into the self-attention of the last six ViT layers (Fig.~\ref{fig:attention_injection}). For hidden states $\mathbf{X} \in \mathbb{R}^{B \times N \times d}$, standard attention first forms queries, keys and values using Eq.~\ref{eq:qkv}.

\begin{equation}\label{eq:qkv}
\mathbf{Q} = \mathbf{X}\mathbf{W}_{q}, \qquad
\mathbf{K} = \mathbf{X}\mathbf{W}_{k}, \qquad
\mathbf{V} = \mathbf{X}\mathbf{W}_{v}
\end{equation}

here, the pretrained projection matrices are denoted $\mathbf{W}_{q}$, $\mathbf{W}_{k}$ and $\mathbf{W}_{v}$, $B$ is the batch size, $N$ the number of tokens and $d$ the hidden width. The guidance is first normalised, $\hat{\mathbf{P}} = \mathbf{P}/\lVert \mathbf{P} \rVert_{2}$, after which layer-specific shifts and a per-sample scalar gate are computed with Eq.~\ref{eq:gate}.

\begin{equation}\label{eq:gate}
\mathbf{g}_{q} = \mathbf{W}_{qg}\,\hat{\mathbf{P}}, \qquad
\mathbf{g}_{k} = \mathbf{W}_{kg}\,\hat{\mathbf{P}}, \qquad
g = \sigma\!\left(\mathbf{w}_{g}^{\top}\hat{\mathbf{P}}\right) \in (0,1)
\end{equation}

here, $\mathbf{W}_{qg}$ and $\mathbf{W}_{kg}$ are learnable guidance projections and $\sigma(\cdot)$ is the logistic sigmoid. The query and key of every token are then shifted according to Eq.~\ref{eq:attninject}.

\begin{equation}\label{eq:attninject}
\mathbf{Q}' = \mathbf{Q} + \alpha\, g\, \mathbf{g}_{q}\mathbf{1}^{\top}, \qquad
\mathbf{K}' = \mathbf{K} + \alpha\, g\, \mathbf{g}_{k}\mathbf{1}^{\top}
\end{equation}

here, $\alpha$ is a learnable scale initialised at $5.0$, and $\mathbf{1}^{\top}$ copies the shift to all $N$ tokens. Attention is then computed in the usual way, as in Eq.~\ref{eq:attn}.

\begin{equation}\label{eq:attn}
\mathrm{Attn}(\mathbf{Q}',\mathbf{K}',\mathbf{V}) = \mathrm{softmax}\!\left(\frac{\mathbf{Q}'\mathbf{K}'^{\top}}{\sqrt{d_{h}}}\right)\mathbf{V}
\end{equation}

here, $d_{h}$ denotes the dimensionality of each attention head. Each modified layer $\ell$ receives its own transformed guidance $\mathbf{P}_{\ell} = \mathbf{W}_{\ell}\,\mathbf{P}$, and these layers are unfrozen so that they can adapt. In this route the guidance changes \emph{how the image is looked at}, redistributing attention over the patches.
\FloatBarrier
\FloatBarrier

\section{Experimental Evaluation}\label{sec:results}

\subsection{Experimental Setup}
Every variant is trained on the same augmented, top-500-filtered OK-VQA corpus, split 80/20 by image identifier so that no photograph appears in both partitions. We optimise with the adaptive moment estimation (Adam) optimiser at a learning rate of $3\times10^{-5}$ and a batch size of 8--16 for at most 15 epochs, stopping early (patience 3--4) once validation exact match no longer improves, and we clip gradient norms at $1.0$ where applicable. The ViT and CLIP encoders stay frozen, apart from the layers explicitly unfrozen for fusion. All runs use one NVIDIA graphics processing unit (GPU) with 8--16\,GB of memory in the Google Colaboratory environment.

Because answers are generated as free text, we report four metrics: exact match (EM) after normalisation; edit similarity, i.e., one minus the normalised Levenshtein distance between prediction and target; token-level F1, the harmonic mean of token precision and recall; and a semantic score that compares prediction and target using contextual embeddings from Bidirectional Encoder Representations from Transformers (BERT), referred to as BERTScore.

\subsection{Comparison of Integration Strategies}
Table~\ref{tab:main} summarises the three variants. Late fusion is best on every metric: it raises EM by $10.0$ points over the unguided baseline and token F1 by $10.4$ points. Attention-matrix injection also beats the baseline, but by a much smaller margin ($+2.4$ EM), and its BERTScore even falls slightly \emph{below} the baseline ($-0.2$). Under this data regime, therefore, the deep injection does not turn semantic guidance into better answers.

\begin{table}
\renewcommand{\arraystretch}{1.3}
\centering
\caption{Exact match (EM), edit similarity (Edit Sim.), token F1 and BERTScore of the three knowledge-integration strategies on the validation split of the augmented OK-VQA subset. The best results are in \textbf{bold}; $\Delta$EM is the gain in EM over the baseline.}
\label{tab:main}
\begin{tabular}{lcccccc}
\hline\noalign{\smallskip}
\textbf{Strategy} & \textbf{Depth} & \textbf{EM (\%)} & \textbf{Edit Sim.} & \textbf{Token F1} & \textbf{BERTScore} & $\Delta$\textbf{EM} \\
\noalign{\smallskip}\hline\noalign{\smallskip}
Baseline (no guidance)   & ---      & 29.44 & 40.57 & 30.97 & 91.21 & ---   \\
Late fusion              & shallow  & \textbf{39.44} & \textbf{48.97} & \textbf{41.34} & \textbf{92.14} & $+10.0$ \\
Attention injection      & deep     & 31.80 & 42.66 & 33.33 & 91.00 & $+2.4$ \\
\noalign{\smallskip}\hline
\end{tabular}
\end{table}

\subsection{Per-Question-Type Analysis}
\FloatBarrier
\begin{figure}[!htbp]
\centering
\includegraphics[width=0.65\textwidth]{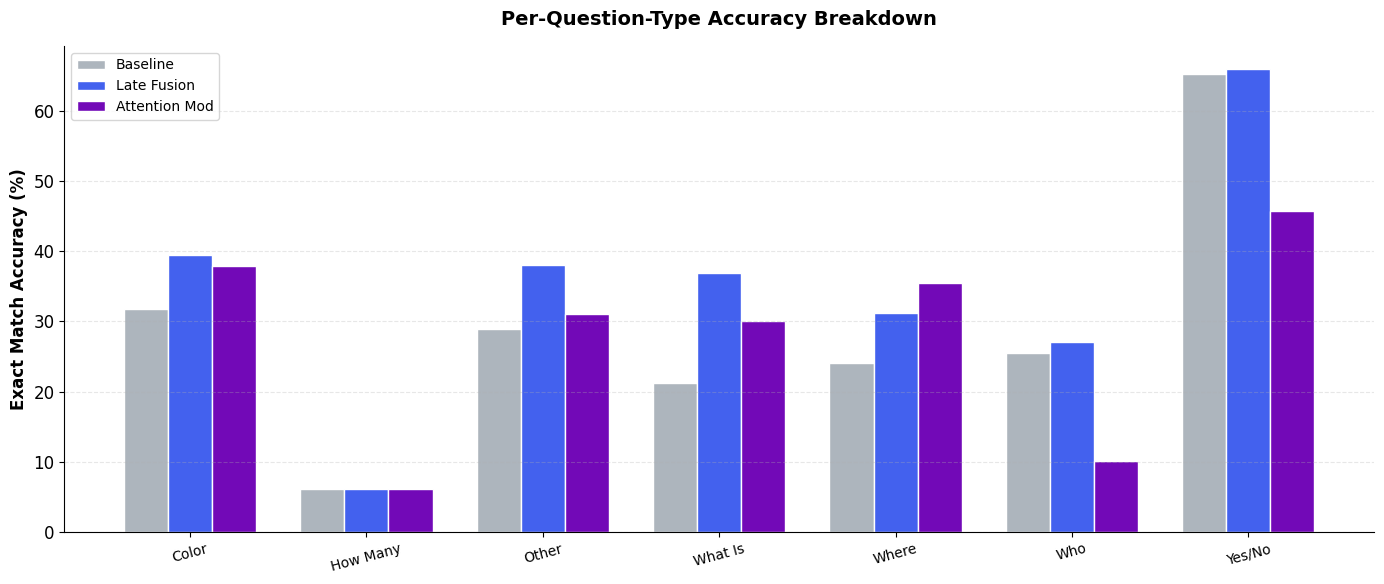}
\caption{Exact match accuracy of the three strategies for each question type.}
\label{fig:qtypes}
\end{figure}
\FloatBarrier
Figure~\ref{fig:qtypes} breaks EM down by question type. Late fusion improves on the baseline in every category, with the largest gains on \emph{What is}, \emph{Other} and \emph{Color} questions, where external context most directly disambiguates the answer. Attention injection is competitive on \emph{Where} questions (35.3, the best of the three) but collapses on \emph{Who} ($10.0$ vs.\ $25.4$ for the baseline) and \emph{Yes/No} ($45.6$ vs.\ $65.2$). Shifting all patch queries and keys along a single guidance direction can therefore disrupt the attention patterns that person-centric and verification-style questions rely on. Counting remains uniformly difficult ($\approx 6\%$), as the guidance concepts carry no numeric signal.

\subsection{Error Distribution}
\begin{figure}[!htbp]
\centering
\includegraphics[width=0.85\textwidth]{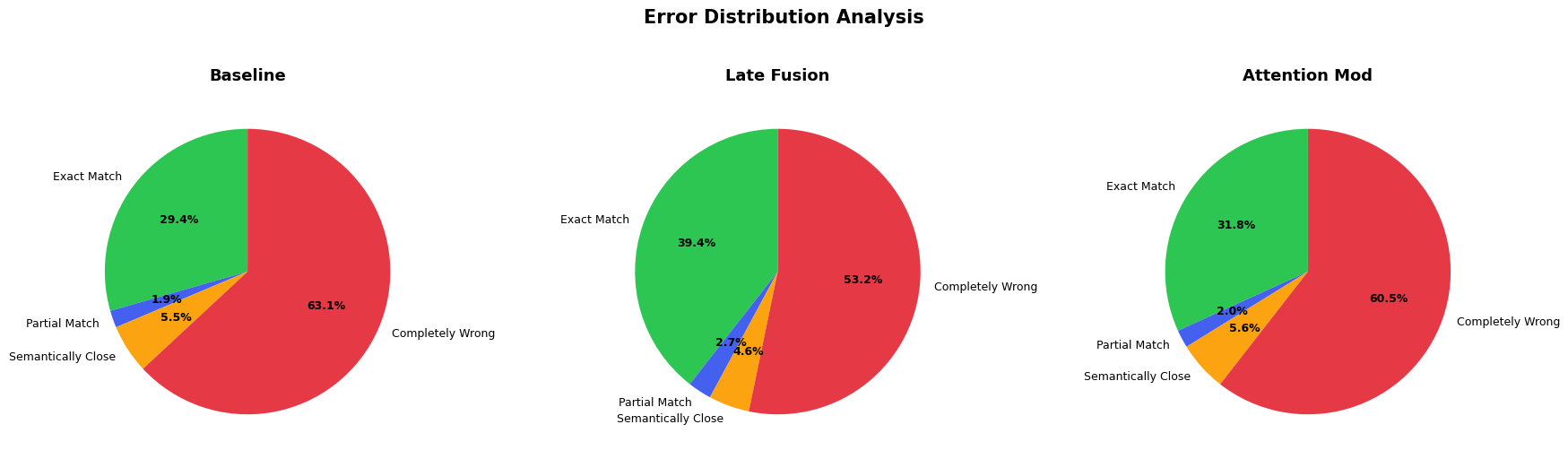}
\caption{Error distribution analysis across the three strategies.}
\label{fig:errors}
\end{figure}

\begin{figure}[!htbp]
\centering
\includegraphics[width=0.70\textwidth]{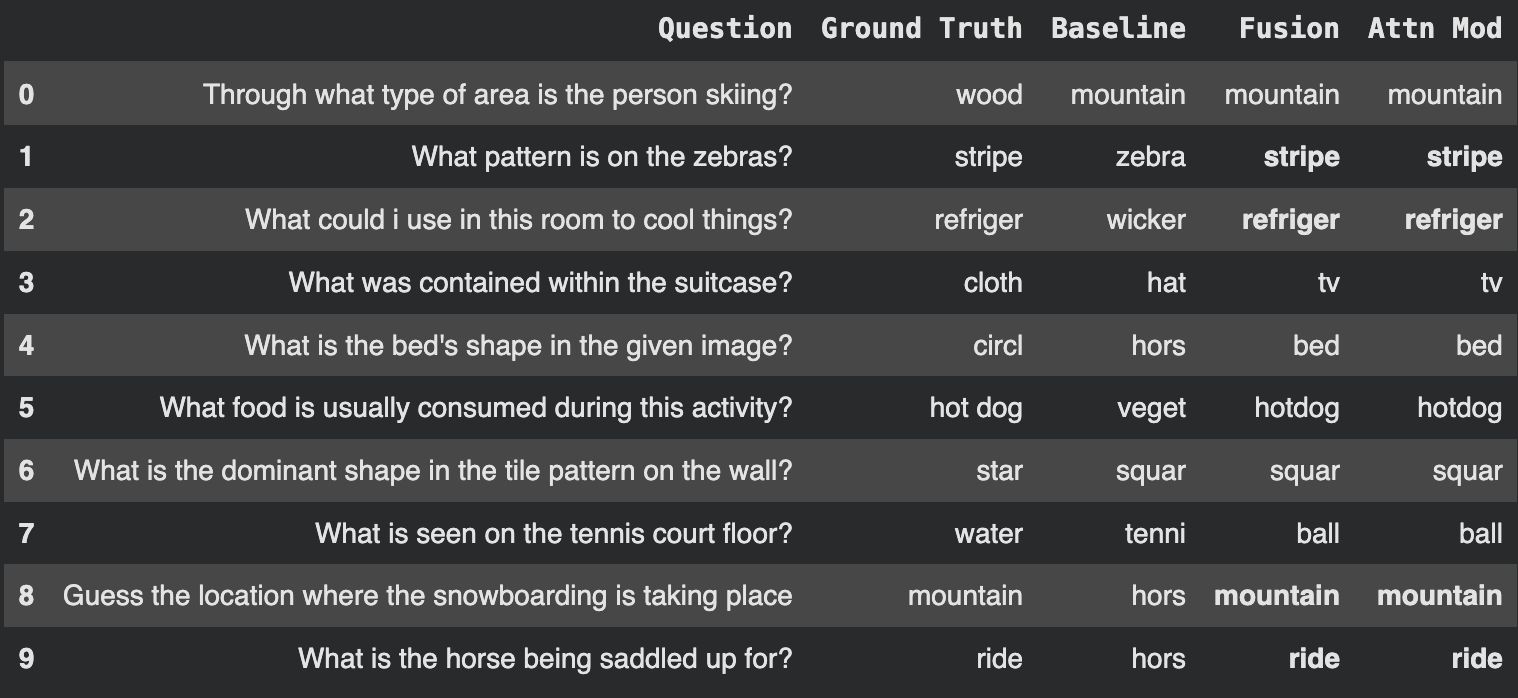}
\caption{Qualitative examples: the guided variants recover attribute- and knowledge-dependent answers that the visual-only baseline misses. Answers are shown after stemming (e.g., \emph{refriger} for \emph{refrigerator}).}
\label{fig:qual}
\end{figure}

Figure~\ref{fig:errors} sorts the predictions into four outcome types. Late fusion lowers the share of completely wrong answers from $63.1\%$ (baseline) to $53.2\%$, whereas attention injection reaches $60.5\%$. The qualitative examples in Fig.~\ref{fig:qual} show both guided models recovering answers that the baseline misses entirely, e.g., \emph{stripe} for the pattern on zebras and \emph{refrigerator} for cooling things. The baseline tends instead to name a salient object rather than the attribute or function asked about, a failure mode that the guidance context corrects.
\FloatBarrier
\FloatBarrier
\FloatBarrier

\subsection{Discussion: Why Does Shallow Beat Deep?}
Our central finding, that late fusion outperforms attention injection by $7.6$ EM points, contradicts the assumption implicit in much recent work~\cite{he2026kvkbvqa,dong2024mail} that deeper knowledge integration is better. We identify three contributing factors.

\emph{(1) Preservation of pretrained representations.} Late fusion leaves the frozen ViT computation untouched. Attention injection perturbs queries and keys in six of the twelve layers (Eq.~\ref{eq:attninject}) and unfreezes those layers, so the model must re-stabilise its visual representation from only about 29k training samples, which is not enough to recover, let alone exceed, the pretrained optimum. This mirrors the negative results on naive attention supervision reported by Samaran et al.~\cite{samaran2021attending}.

\emph{(2) Granularity mismatch.} The guidance vector $\mathbf{P}$ is a single global direction (a mean over concept embeddings). Injecting one global direction into the query and key of \emph{every} patch token (the $\mathbf{1}^{\top}$ broadcast in Eq.~\ref{eq:attninject}) cannot express \emph{which region} each concept refers to; it biases attention uniformly, which helps some question types (\emph{Where}) and harms others (\emph{Who}, \emph{Yes/No}). Late fusion treats $\mathbf{P}$ as what it actually is, a global semantic summary, and fuses it at the level of the global representation.

\emph{(3) Optimisation burden.} Attention injection introduces per-layer projections, gates and scaling factors that must be learned jointly with unfrozen ViT layers, whereas late fusion trains a single linear fusion layer. In the 6 samples-per-class regime the smaller hypothesis space wins, consistent with fusion-depth findings in adjacent multimodal domains~\cite{tziafas2022fusion}.

These results do not imply that deep integration is inherently inferior, but rather that \emph{integration depth must be matched to the data regime}: under scarce fine-tuning data with frozen backbones, decision-level fusion of LLM guidance is the more reliable choice.

\section{Conclusion}\label{sec:con}
This paper presented an LLM-guided VQA framework in which structured reasoning (visual focus concepts, distractor elements and external knowledge) is generated by Llama~3.3~70B, encoded with CLIP, and integrated into a model built from a frozen ViT and FLAN-T5, together with a paraphrasing pipeline that expands the training data more than tenfold. A systematic comparison under identical data-scarce conditions shows that shallow, decision-level late fusion delivers the largest gains ($+10.0$ EM), while deep attention-matrix injection yields only a marginal improvement and can disrupt the attention patterns that certain question types require. We attribute this to the preservation of pretrained representations, the global granularity of the guidance signal, and the optimisation burden under scarce data.

Future work may inject concept-level guidance with spatial correspondence, giving deep integration the regional signal it currently lacks; repeat the comparison at larger data scales to locate the crossover between shallow and deep integration; and extend the guidance schema with numeric fields for counting questions, with evaluation on the official OK-VQA protocol for leaderboard comparability.

%
%
\bibliographystyle{unsrt}
\bibliography{refupdated}

\end{document}